\documentclass[runningheads]{llncs}

\usepackage{eccv}

\usepackage{eccvabbrv}

\usepackage{graphicx}
\usepackage{booktabs}

\usepackage[accsupp]{axessibility}  % Improves PDF readability for those with disabilities.

\usepackage{graphicx}
\usepackage{amsmath}
\usepackage{amssymb}
\usepackage{booktabs}
\usepackage{stfloats} % image arrange
\usepackage{xspace} % space
\usepackage{multirow}
\usepackage{tabularray}
\usepackage{bbding}
\usepackage{pifont}
\usepackage{mathrsfs}

\usepackage[pagebackref,breaklinks,colorlinks,citecolor=eccvblue]{hyperref}
\usepackage{orcidlink}
\definecolor{dgreen}{rgb}{0.0,0.6,0.0}
\newcommand{\cmark}{\textcolor{dgreen}{\ding{51}}}
\newcommand{\xmark}{\textcolor{red}{\ding{55}}}

\begin{document}

\title{Masked Video and Body-worn IMU Autoencoder for Egocentric Action Recognition}

\titlerunning{Masked Video and Body-worn IMU Autoencoder}

% \author{Mingfang Zhang\inst{1}\orcidlink{0000-1111-2222-3333} \and
% Second Author\inst{2,3}\orcidlink{1111-2222-3333-4444} \and
% Third Author\inst{3}\orcidlink{2222--3333-4444-5555}}

\author{Mingfang Zhang \and
Yifei Huang$^*$ \and
Ruicong Liu \and
Yoichi Sato}

\authorrunning{M.~Zhang et al.}

\institute{Institute of Industrial Science, the University of Tokyo, Tokyo, Japan
\\
\email{\{mfzhang,hyf,lruicong,ysato\}@iis.u-tokyo.ac.jp}}

\maketitle
\let\thefootnote\relax\footnotetext{ $^*$Corresponding author.}
\begin{abstract}
    Compared with visual signals, Inertial Measurement Units (IMUs) placed on human limbs can capture accurate motion signals while being robust to lighting variation and occlusion. While these characteristics are intuitively valuable to help egocentric action recognition, the potential of IMUs remains under-explored. In this work, we present a novel method for action recognition that integrates motion data from body-worn IMUs with egocentric video. Due to the scarcity of labeled multimodal data, we design an MAE-based self-supervised pretraining method, obtaining strong multi-modal representations via modeling the natural correlation between visual and motion signals. To model the complex relation of multiple IMU devices placed across the body, we exploit the collaborative dynamics in multiple IMU devices and propose to embed the relative motion features of human joints into a graph structure. Experiments show our method can achieve state-of-the-art performance on multiple public datasets. The effectiveness of our MAE-based pretraining and graph-based IMU modeling are further validated by experiments in more challenging scenarios, including partially missing IMU devices and video quality corruption, promoting more flexible usages in the real world.

    \keywords{Egocentric action recognition \and Inertial Measurement Units \and Multimodal Masked Autoencoder }
\end{abstract}

\section{Introduction}
\label{sec:intro}

\begin{figure}[t]
  \centering
   \includegraphics[width=1.0\linewidth]{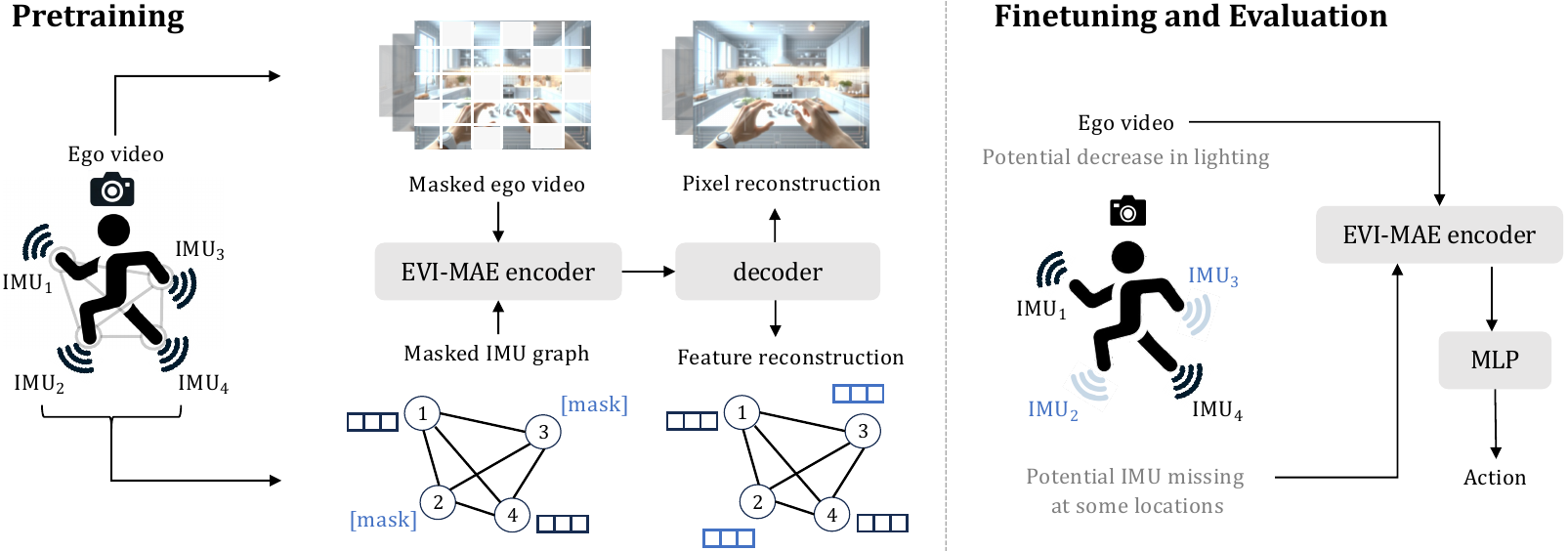}
   % \vspace{-2em}
   \caption{
   Overview of our EgoVideoIMU-MAE (EVI-MAE). Because of the scarcity of labeled multimodal data, we propose an MAE-based pretraining approach with unlabeled egocentric video and IMU signals. To exploit the collaborative dynamics in multiple IMU devices, we propose to embed the relative motion features of human joints into a graph. In the finetuning and evaluation phase, we consider potential video and IMU corruption for more flexible usages.
   % In the pretraining phase, we construct a graph with IMU features and mask some of the feature nodes in the graph and pixel patches in the ego video. The encoder operates on visible nodes and patches, and the decoder tries to reconstruct the original input. 
   % In the finetuning and evaluation phase, we propose a challenge of IMU device missing for more flexible usages. Our EM-MAE takes ego video as optional input. It can work with solely IMU modality for user's privacy concerns.
   }
   \label{fig:teasor}
   % \vspace{-1.5em}
\end{figure}

Advances in wearable cameras and egocentric video datasets bring egocentric action recognition as one pivotal aspect of understanding human behavior~\cite{egoinstructor,somasundaram2023project,huang2018predicting}. 
Because of the inherent difficulty in recognizing actions solely from egocentric videos, recent successes embrace multimodal signals, integrating vision, language, and audio to enrich the contextual understanding of actions~\cite{gong2023mmg, kazakos2019epic}. 
However, one affordable alternative, an Inertial Measurement Unit (IMU) that captures acceleration, angular velocity, and orientation of human limbs, remains under-explored. In addition to being cheap and energy-efficient, when affixed to human limbs, IMUs can accurately record 3D human movements which are often hard to observe because of the viewpoint limitation of egocentric cameras~\cite{huang2020improving}. IMUs excel at motion sensing but do not contain visual information~\cite{nakamura2021sensor}. On the other hand, videos provide visual information but are vulnerable to harsh imaging conditions and not optimal for sensing motion~\cite{chen2021darklight, zhang2021optical}. Therefore, joint modeling of video and IMU is a promising direction to leverage their complementarity. 
% The use of IMUs in egocentric action recognition holds great potential ranging from augmented reality~\cite{cutolo2020ambiguity,kabuye2023mixed} to healthcare~\cite{pesenti2023imu,routhier2020clinicians}.

% ranging from AR and VR to healthcare and sports science.

% 1 challenge:      each single modality is vulnerable
% 1 observation:    modality collaboration
% 1 solution:       multimodal fusion -- too naive, ignore this point

% 2 challenge:      dataset scale limitation
% 2 observation:    modality correlation 
% 2 solution:       multimodal pretraining

% 3 challenge:      IMU device relation understanding
% 3 observation:    IMU devices are collaborative
% 3 solution:       use graph

% 1 challenge:      each single modality is vulnerable
% If action recognition depends solely on IMUs, the primary challenge is their inability to recognize specific objects and contextual details in a scene. For instance, distinguishing between actions such as `washing hands' and `washing tomatoes' becomes challenging when only motion data are analyzed. However, compared to the video modality, IMU signals offer the advantage of preserving user privacy while reducing the need for computational power and storage capacity. Additionally, IMUs exhibit inherent resilience to common visual modality challenges such as occlusions, varying perspectives, changes in lighting, and background distractions. 
% This resilience makes IMUs an invaluable tool for analyzing human actions.

However, two significant challenges prevent us from directly applying existing multimodal action recognition frameworks~\cite{gong2022contrastive,georgescu2023audiovisual} to this task. The first one is the scarcity of data. While several recent egocentric datasets~\cite{ego4d,gong2023mmg} offer large-scale multimodal egocentric data including IMU, they are measured by camera IMUs and thus can only provide motion clues that can also be inferred from the video. 
Very few datasets~\cite{de2009guide,bock2023wear} contain synchronized video and separately attached IMU sensor data, but the largest dataset~\cite{bock2023wear} offers only 9 hours of footage with action annotations. 
The other challenge lies in the complexity of multiple IMU devices across the body.
Since each IMU device can provide motion signals of only a single body part. However, a holistic understanding of motions of different body parts is necessary for recognizing various actions. Consequently, modeling these intricate multi-device relations becomes challenging.

% 1 observation:    modality collaboration
% 1 solution:       multimodal fusion
% Regarding the first challenge, although each single modality is vulnerable, we assert that \textit{egocentric video and IMU signals are complimentary}. Egocentric video captures the dynamic appearance of objects and surroundings, yet it contains significant temporal redundancy and correlation []. In contrast, IMU signals directly focus on the subject's movements, closely aligning with the actions' verbs, but they fall short in providing insights into the subject's surrounding context.

% 2 observation:    modality correlation 
% 2 solution:       multimodal pretraining
In this work, we propose a novel multi-modal egocentric action recognition method using both egocentric video and body-worn IMU signals. 
To address the challenge of data scarcity, we observe that \textbf{egocentric video and IMU signals are correlated}: 1), in egocentric videos, the movements of visible upper limbs show a strong correlation with the IMU signals attached to these limbs. 2), the global motion in the camera view is closely related to the IMU signals from the subject's lower limbs. To exploit this correlation, we develop a multimodal Masked Autoencoder (MAE)~\cite{he2022masked} based pretraining method. Specifically, we use a joint model to take masked video and masked IMU signals as inputs and reconstruct the two modalities simultaneously. With this self-supervised learning method, we can leverage the large volume of unlabeled video-IMU data to learn the natural alignment and the complementary aspect between the two modalities. This allows us to leverage larger-scale data without manual action annotation, which is a major advantage of our method compared to existing fully supervised methods.
Regarding the second challenge, we observe that \textbf{multiple IMU devices are collaborative}. IMU devices positioned at various locations on the human body collect complementary data, providing a holistic picture of body movements. These IMU sensors on human limbs reflect the \textit{relative motion} of human joints.
Drawing inspiration from skeleton-based action recognition models \cite{yan2018spatial, du2015hierarchical}, we propose a graph-based IMU modeling technique, where we construct a graph with nodes as IMU features. This graph structure can effectively capture motion relationships among different IMU sensors, and the graph neural network can therefore optimize the representation learning. 

An overview of our proposed method can be found in Figure~\ref{fig:teasor}. During the MAE pretraining, some of the IMU feature nodes are masked together with egocentric video patches. Our EgoVideoIMU-MAE (EVI-MAE) encoder operates on visible nodes and patches, and the decoder tries to reconstruct the original input. In this way, we obtain an encoder with knowledge of intra-IMU relationships and the video-IMU correlations. The representations learned by this multimodal pretraining technique contain the complementarity of the two modalities and thus can be robust to less restricted scenarios where one modality becomes unreliable, \textit{e.g.}, video quality decline, or partial IMU devices missing.

% Real world application
Experiments on two public datasets~\cite{de2009guide,bock2023wear} show that the representation learned by our EVI-MAE significantly outperforms the performance of baseline models without pretraining or with single-modal pretraining. Our ablation studies confirm the efficacy of constructing the IMU feature graph, regardless of whether pretraining is involved.
Considering constraints in real-world applications, we further test our model by designing experiments in various challenging scenarios: 1) partially missing IMU devices, 2) video quality variation, and 3) cross-dataset generalization. Extensive results in these challenging scenarios demonstrate the robustness and adaptability of our model.
% Since the number and placement of IMUs may vary across different users in different environments, 
% potentially leading to discrepancies compared with the IMU settings in our training data. We assume to have all the IMU devices for pretraining, and address the challenge of potential device missing in the finetuning and evaluation. In the pretraining phase, we mask some of the node features to simulate the IMU device missing scenario. Because the model is trained to reconstruct the whole IMU graph, it learns to infer global motion patterns from partial data. Regarding the video modality, the visual appearance is significantly affected by ambient lighting, making visual models vulnerable to variations in lighting conditions. Therefore, we employ a sophisticated technique [] to synthesize low-light effects and reduce the signal-to-noise ratio on videos and evaluate our multimodal method. Our method gains an advantage from using IMU features unaffected by visual appearance, leading it to depend more on IMU data in visually challenging scenarios.

% Experiment results
% We also conduct tests under challenging conditions, including finetuning with missing devices, cross-dataset finetuning, and evaluations using synthetic low-light videos. These tests further demonstrate the robustness and adaptability of our model.

% Contributions
In summary, our main contributions are as follows:
\begin{itemize}
% \vspace{-0.8em}
\item{We design the first method for joint egocentric video and body-worn IMU representation learning by multimodal masked autoencoding.}
% \vspace{-0.8em}
\item{We propose a graph-based IMU modeling technique to better leverage the collaborative nature of IMU devices on different body locations.}
% \vspace{-0.8em}
\item {
Experiments show that our method outperforms various baseline methods and continues to exhibit robust performance and adaptability in challenging scenarios, promoting more flexible usages in the real world.}
\end{itemize}  
\section{Related Works}

\paragraph{IMU-based action recognition} 
IMU-based action recognition has seen significant progress.
The DeepConvLSTM \cite{ordonez2016deep} model merges convolutional and recurrent layers to extract features while modeling temporal dependencies. Subsequent efforts \cite{murahari2018attention, yao2017deepsense} have expanded upon this model, introducing novel architectures \cite{abedin2021attend, xu2019innohar} that further advance the field. Nevertheless, these advancements tend to overlook the intricacies of inter-device relationships among multiple IMUs.
SADeepSense \cite{yao2019sadeepsense} and subsequent studies \cite{ma2019attnsense, liu2020giobalfusion} incorporates the attention mechanism \cite{vaswani2017attention} to facilitate the fusion of data from heterogeneous sensors, such as accelerometers and gyroscopes, positioned across the body. However, the limited size of datasets with labeled actions constrains their performance.
Recent researches \cite{yuan2022self, tang2020exploring} introduce the concept of contrastive learning to IMU representation learning. LIMU-BERT \cite{xu2021limu} adapts BERT \cite{devlin2018bert} for IMU signals, which involves randomly masking a small portion of data and reconstructing them.
Compared to previous IMU-based action recognition methods, we propose a representation learning method that utilizes unlabeled IMU signals and also learns multiple IMU device relationships by proposing to construct a graph with IMU features. Moreover, our method incorporates egocentric videos to exploit multimodal relations.

\paragraph{Egocentric multimodal action recognition} 
Most previous action recognition methods are conducted on videos \cite{feichtenhofer2019slowfast, bertasius2021space, wang2018appearance, tateno2024learning}. However, for egocentric action recognition, where the videos often have limited field-of-view and large motion blur \cite{liu2024single}, incorporating complementary multimodal information becomes crucial~\cite{huang2020mutual}.
Recent advancements have seen the emergence of many egocentric video datasets \cite{somasundaram2023project, ego4d, grauman2023ego, damen2022rescaling, sener2022assembly101, zhang2022egobody, liu2022hoi4d, huang2024egoexolearn} that encompass diverse modalities, including video, depth, audio, language, and motion sensors.
Prior approaches to multimodal fusion in egocentric activity recognition range from straightforward concatenation \cite{poria2016convolutional, xiao2020audiovisual} to more sophisticated tensor decomposition techniques \cite{liu2018efficient}. MMG \cite{gong2023mmg} aims at addressing the challenge of zero-shot modality mismatch challenge.
However, the majority of existing datasets and methods focus on sensing devices attached to the subject's head, primarily through head-mounted cameras \cite{sudhakaran2019lsta, wang2023ego}, microphones \cite{kazakos2019epic, yang2022interact}, and IMUs \cite{tsutsui2021you, gong2023mmg}, while neglecting sensors placed on the human limbs. In contrast to existing works, we propose an egocentric multimodal representation learning approach that integrates visual signals from head-mounted devices with motion signals from limb-mounted sensors.

\paragraph{Multimodal masked autoencoders}

% The use of visual modalities for pretraining and action recognition [1,2,3,4] is widely practiced. In our study, we introduce a Masked Autoencoder tailored for IMU sensors attached to human limbs. 

% Recent multimodal pretraining methods achieves significant strides and the Masked Autoencoder (MAE) stands out for its ability to integrate and learn from different modalities, such as semantically related image and language [1,2], and synchronized audio and visual data [3,4]. MAE learns a meaningful representation with the pretext task of recovering the original inputs or features from the corrupted ones. With multimodal input, MAE learns a joint representation that fuses the unimodal signals and reconstruct each modality with the joint representation.

The concept of data masking has been extensively studied since \cite{vincent2008extracting}. Masked Autoencoder (MAE) \cite{he2022masked} distinguishes itself by its proficiency in learning meaningful representations. Its applicability spans across various domains, including image processing \cite{he2022masked}, video analysis \cite{tong2022videomae, feichtenhofer2022masked}, audio processing \cite{huang2022masked}, point cloud data interpretation \cite{pang2022masked}, and graph analysis \cite{hou2022graphmae}. Additionally, MAE demonstrates robust capabilities in integrating and learning from multiple modalities. This includes image and language data \cite{geng2022multimodal, kwon2022masked}, as well as synchronized audio, images \cite{gong2022contrastive} and videos \cite{georgescu2023audiovisual}, showcasing its effectiveness in handling different data representations together. This paper introduces an innovative approach utilizing an MAE-based method to leverage the synergistic and complementary interactions between egocentric video and body-worn IMU motion data for effective action recognition. 
\section{Method}

% \subsection{Masked Autoencoders Preliminaries}

% Our method is designed based on the operational logic of ImageMAE [], in which the input image $D_{im} \in \mathbb{R}^{H \times W \times 3}$ is tokenized to $T_{im}$ in accordance with the setup of prior supervised learning methods using transformers, by conducting a linear projection on the image to get non-overlapping patches. 
% Positional embeddings $P_{im}$ are then appended to each patch. 
% Subsequently, a random subset of these tokens $TM_{im}$ is masked, and only the unmasked tokens $TU_{im}$ are forwarded to the transformer encoder with learnable mask tokens. 
% In the final step, a transformer decoder is employed to attempt the reconstruction of the original image $D_{im}$ in pixel space, utilizing the mean squared error (MSE) as the loss function.

\subsection{Problem definition}

We use synchronized video and IMU signals of temporal length $T$ as inputs for pretraining and finetuning the model. The video can be represented as $\boldsymbol{D}_v \in \mathbb{R}^{T \times \mathcal{S}_v \times H \times W \times 3}$, where $\mathcal{S}_v$ is the video sampling rate, and $H, W$ represent the height and width of the video, respectively. For the IMU signal, we consider $N_{imu}$ IMU sensors installed on the person's limbs. For each IMU, we read the acceleration in the $x$, $y$, and $z$ axes while discarding other readings like angular velocity and orientation because acceleration is universally presented across most types of IMUs. Even with the need for including the discarded readings, the proposed method can be easily extended. Overall, the raw IMU signals can be represented as $\boldsymbol{D}_{raw} \in \mathbb{R}^{N_{imu} \times T \times \mathcal{S}_{imu} \times 3}$, where $\mathcal{S}_{imu}$ is the sampling rate of IMU. Our goal for pretraining is to learn a multimodal representation effective for the downstream action recognition task, which is a classification problem given predefined $C$ classes.

% We also consider challenges in real-world scenarios, including finetuning with a less number of IMU devices and evaluation with low-light videos. We aim to design a model to address these challenges without modifying the model structure. The experiment settings are detailed in Section 4.x.

\begin{figure}[t]
  \centering
  % \fbox{\rule{0pt}{1.2in} \rule{1.0\linewidth}{0pt}}
   \includegraphics[width=1.0\linewidth]{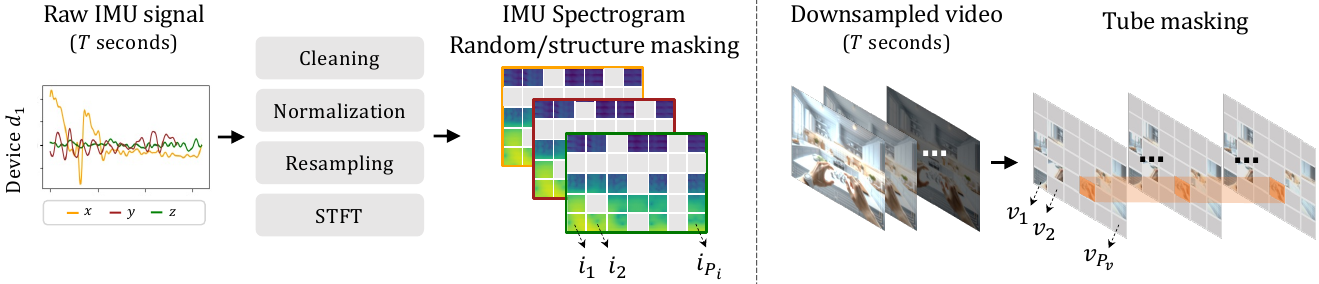}
   % \vspace{-2em}
   \caption{
   IMU and video data preprocessing and masking.
   }
   \label{fig:patch_mask}
   % \vspace{-1.5em}
\end{figure}

\subsection{Data preprocessing and masking}

% Patch
As shown in \cref{fig:patch_mask}, the preprocessing of the $T$-second raw IMU signal involves several steps.
Due to varying sample rates across IMU sensors, signals are resampled to a consistent rate and normalized to a uniform value range. The STFT (Short Time Fourier Transform) step then processes each 1-dimensional acceleration signal, consisting of $T_{imu}$ resampled temporal points, converting it into a sequence of features of $M_{imu}$ dimensions as a spectrogram $\boldsymbol{D}_{spec} \in \mathbb{R}^{T_{imu} \times M_{imu}}$. This spectrogram is subsequently divided into $P_{imu}$ square patches, $\boldsymbol{i} = \left[i_1, i_2, ..., i_{P_{imu}}\right]$, which serve as the input for the model. For the egocentric video, each $T$-second video segment is temporally down-sampled to $T_v$ frames and then divided into $P_v$ square patches $\boldsymbol{v} = [v_1, v_2, ..., v_{P_v}]$.

% Mask
For the IMU spectrograms, we employ unstructured random masking with a masking ratio of $R_{imu}$ that transforms IMU spectrogram patches $\boldsymbol{i}$ into $\boldsymbol{i}_m$.
% $\boldsymbol{i}_m = \psi_{i}(\boldsymbol{i}, R_{imu})$. 
We also report the performance of structured masking in \cref{tab:ablation_main}, including randomly masking a portion of time and frequency of $\boldsymbol{i}$. For videos, we employ the tube masking strategy \cite{tong2022videomae} with a high masking ratio of $R_{v}$, where the masking locations are the same for all frames, which transforms video patches $\boldsymbol{v}$ into $\boldsymbol{v}_m$.
% $\boldsymbol{v}_m = \psi_{v}(\boldsymbol{v}, R_{v})$.

\subsection{IMU feature graph}
\label{sec:IMU_graph}

To model the complex relationships among multiple IMU devices positioned across the body, we exploit the collaborative dynamics of these devices and propose to embed the relative motion features of human joints into a graph structure. 
% Specifically, we begin by encoding all the IMU tokens, $\boldsymbol{i}$, using an IMU encoder. Note that $\boldsymbol{i}$ is from all $N_{imu}$ IMU devices before any masking process. The encoded IMU features from each device can be represented as $F_{d_1}, F_{d_2}, ... F_{d_{N_{imu}}}$.
Specifically, we initiate the process by encoding all the IMU spectrogram patches $\boldsymbol{i}$ via an IMU encoder. In particular, $\boldsymbol{i}$ encompasses data from all the $N_{imu}$ IMU devices, $d_1, d_2, ..., d_{N_{imu}}$, prior to the application of any masking procedures. The encoded features from each IMU device are denoted as 
$ \boldsymbol{f}_{d} = [\boldsymbol{f}_{d}^{1}, \boldsymbol{f}_{d}^{2}, \ldots, \boldsymbol{f}_{d}^{{N_{imu}}}]$.

Subsequently, we construct an IMU feature graph $\mathcal{G} = (\mathcal{V},\boldsymbol{A},\boldsymbol{f}_{d})$, where $\mathcal{V} = \{\nu_n\}_{n=1}^{N_{imu}}$ denotes the set of nodes corresponding to the IMUs distributed across the body, $\boldsymbol{A} \in \{0,1\}^{N_{imu}\times N_{imu}}$ represents the adjacency matrix, and $\boldsymbol{f}_{d} \in \mathbb{R}^{N_{imu} \times M_{imu}}$ is IMU node features. 
The adjacency matrix $\boldsymbol{A}$ is designed to be fully connected, highlighting that actions usually require coordinated movements across different limbs. The efficiency of this graph-based modeling is guaranteed since in practical applications, individuals typically do not wear a large number of IMU devices.

\subsection{Pretraining pipeline}

\begin{figure}[t]
  \centering
  % \fbox{\rule{0pt}{1.8in} \rule{1.0\linewidth}{0pt}}
   \includegraphics[width=\linewidth]{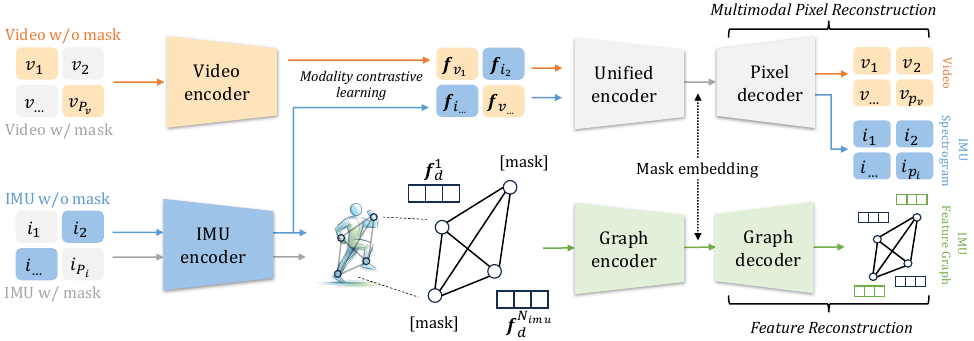}
   % \vspace{-2em}
   \caption{
   Our EVI-MAE pretraining network processes video patches $\boldsymbol{v}$ and IMU spectrogram patches $\boldsymbol{i}$ and incorporates them into two branches, a multimodal pixel reconstruction branch, and an IMU feature reconstruction branch.
   }
   \label{fig:network}
   % \vspace{-1.5em}
\end{figure}

Because of the scarcity of labeled multimodal data, we design an MAE-based self-supervised pretraining method. We aim to effectively capture strong multimodal representations by leveraging the inherent correlation between visual and motion signals. As shown in \cref{fig:network}, the pretraining network structure incorporates two primary branches, the multimodal pixel reconstruction branch and the IMU feature reconstruction branch.

\vspace{0.5em}
\noindent \textbf{3.4.1 Multimodal pixel reconstruction branch}
\vspace{0.2em}

\noindent As shown in \cref{fig:network}, this branch involves a video encoder, an IMU encoder, a unified encoder, and a pixel decoder.
In this branch, we use the visible patches from a masked video, $\boldsymbol{v}_m$, and the visible patches from masked IMU spectrograms, $\boldsymbol{i}_m$, as inputs. Our objective is to reconstruct the masked video and the masked IMU spectrograms. By doing so, we aim to exploit the abundant supply of unlabeled yet synchronized video-IMU data to understand the natural alignment and complementary interaction between these two modalities, thereby achieving a strong multimodal representation.

Specifically, $\boldsymbol{v}_m$ and $\boldsymbol{i}_m$ are first incorporated with positional embeddings $\boldsymbol{p}_{v}, \boldsymbol{p}_{imu}$ and modality type embeddings $\boldsymbol{m}_{v}, \boldsymbol{m}_{imu}$. The positional embedding encodes the spatial positions of patches to maintain their spatial relationships and the modality type embedding distinguishes between video and IMU modality.
Then, they are inputted into the video encoder and the IMU encoder, respectively. 
The video encoder utilizes a standard ViT backbone \cite{dosovitskiy2020vit} and incorporates joint space-time attention \cite{arnab2021vivit, liu2022video}, allowing tokens to interact within the multi-head self-attention layer. 
Meanwhile, the IMU encoder is designed following ImageMAE \cite{he2022masked}, since we convert the 1D motion signals to 2D spectrogram images. 
As shown in \cref{fig:network}, the visible video and IMU encoded features, denoted as $\boldsymbol{f}_{v}$ and $\boldsymbol{f}_{i}$, are then merged into a unified embedding and processed by a modality-unified encoder, concluding the encoding phase.

For decoding, this multimodal embedding is first padded with trainable masked tokens at the masked positions as $\boldsymbol{e}_x = [\boldsymbol{e}_{imu}, \boldsymbol{e}_v]$. Along with the addition of both modality type embeddings $\boldsymbol{m}'_{imu}, \boldsymbol{m}'_{v}$, and positional embeddings, $\boldsymbol{p}'_{imu}, \boldsymbol{p}'_{v}$, they are passed through a transformer-based pixel decoder, $\mathcal{D}_{uni}$, \ie, 
\begin{equation}
    O_p = \mathcal{D}_{uni}([\boldsymbol{e}_{imu}, \boldsymbol{e}_v] + [\boldsymbol{m}'_{imu}, \boldsymbol{m}'_{v}] + [\boldsymbol{p}'_{imu}, \boldsymbol{p}'_{v}]).
\end{equation}
Finally, we partition the output pixels $O_p$ as video and IMU spectrograms and rearrange the pixel patches to align with the original $\boldsymbol{v}$ and $\boldsymbol{i}$.

\vspace{0.5em}
\noindent \textbf{3.4.2 IMU feature reconstruction branch}
\vspace{0.2em}

\noindent As shown in \cref{fig:network}, this branch involves the same IMU encoder, a graph encoder, and a graph decoder.
In this branch, we shift our focus from patch level to IMU device level. Based on the structure of IMU feature graph, we design an MAE-based pretraining method that uses a corrupted graph with masked nodes to reconstruct the original graph. In this way, we aim to capture the intricate relationships among different IMU sensor features, thereby enhancing the model's comprehension of motion dynamics and sensor interactions from a holistic perspective. 

We first construct an IMU feature graph $\mathcal{G} = (\mathcal{V},\boldsymbol{A},\boldsymbol{f}_{d})$ as described in \cref{sec:IMU_graph}.
Inspired by GraphMAE \cite{hou2022graphmae}, our pretraining output is $\mathcal{\hat{G}}$, whose goal is to reconstruct $\mathcal{G}$ from a corrupted input feature $\boldsymbol{f}_{dc}$ as:

\begin{equation}
    \boldsymbol{f}_g = GraphEnc(\boldsymbol{A},\boldsymbol{f}_{dc}),\ \mathcal{\hat{G}} = GraphDec(\boldsymbol{A},\boldsymbol{f}_g).
\end{equation}

Specifically, we employ a masking procedure where a ratio, $R_g$, of the $N_{imu}$ node features in $\boldsymbol{f}_{d}$ are selectively masked by replacing them with mask tokens, resulting in a corrupted graph with features, $\boldsymbol{f}_{dc}$. This corrupted graph is then processed using a Graph Isomorphism Network (GIN) \cite{xu2018powerful} encoder, chosen for its outstanding inductive bias suited for graph-level classification tasks. 
% On the other hand, we find that the Graph Attention Network [] exhibit unstable performance in our task because it is more expressive in node classification []. 
During the decoding phase, the encoded graph features $\boldsymbol{f}_g$, which shares the same shape with $\boldsymbol{f}_{d}$, are once again substituted with learnable mask embeddings at the previously masked positions. At this stage, we reintroduce the GIN network to serve as the decoder to reconstruct the entire graph. 
% The output is denoted as $\mathcal{G'}$.

\vspace{0.5em}
\noindent \textbf{3.4.3 Pretraining criterion}
\vspace{0.2em}

\noindent For pixel reconstruction, we calculate the mean squared error (MSE) between the pixels of the reconstructed video and those of the original video, as well as the pixels of the IMU spectrogram. In the case of IMU feature graph reconstruction, we use the cosine similarity error to compare the reconstructed IMU device-level features $\boldsymbol{\hat{f}}_d$ and the original $\boldsymbol{f}_d$, within the set of masked node features $\mathcal{V}_c$. The cosine similarity error is defined as follows:
\begin{equation}
    L_{cos} = \frac{1}{|\mathcal{V}_c|}\sum_{\nu_n \in \mathcal{V}_c} (1-\frac{(\boldsymbol{\hat{f}}_{d}^{\nu_n})^T\boldsymbol{f}_{d}^{\nu_n}}{||\boldsymbol{\hat{f}}_{d}^{\nu_n}||\cdot||\boldsymbol{f}_{d}^{\nu_n}||}),
\end{equation}
where $\boldsymbol{f}_{d}^{\nu_n}$ denotes the IMU feature of device on node $\nu_n$.
Inspired by a multimodal MAE method \cite{gong2022contrastive}, we further enrich our approach by integrating a contrastive objective, leveraging the output features $\boldsymbol{f}_v,\boldsymbol{f}_i$ from both video and IMU encoders:

% \iffalse
\begin{equation}
\label{equ:contrast}
\resizebox{.9\hsize}{!}{$L_{con} = -\frac{1}{2N_{b}} \big(\sum_{k=1}^{N_{b}}\text{log}\frac{\text{exp}(s(\boldsymbol{f}_v^k,\boldsymbol{f}_i^k)/\tau)}{\sum_{j=1}^{N_{b}}\text{exp}(s(\boldsymbol{f}_v^k,\boldsymbol{f}_i^j)/\tau)} + \sum_{k=1}^{N_{b}}\text{log}\frac{\text{exp}(s(\boldsymbol{f}_v^k,\boldsymbol{f}_i^k)/\tau)}{\sum_{j=1}^{N_{b}}\text{exp}(s(\boldsymbol{f}_v^j,\boldsymbol{f}_i^k)/\tau)}\big),$}
\end{equation}
% \fi
where $N_{b}$ is the batch size, $s(\boldsymbol{f}_v^k,\boldsymbol{f}_i^k) = (\boldsymbol{f}_v^k)^T(\boldsymbol{f}_i^k)$, and $\tau$ is the temperature.
By utilizing supervision formed through non-corresponding video and IMU pairs within the same batch, our model is adept at learning features that are more aligned across modalities. In summary, our total loss is $L = \alpha L_{mse} + \beta L_{cos} + \gamma L_{con}$.

\subsection{Finetuning for action recognition}

In the fine-tuning stage, we discard the pixel and graph decoders and fine-tune only the encoders. 
This process involves using all video and IMU patches, denoted as $\boldsymbol{v}$ and $\boldsymbol{i}$.
We then concatenate the output features of the unified encoder and the graph encoder, apply global average pooling, and then use a linear classifier on top for the action classiﬁcation tasks.

\section{Experiments}

\subsection{Datasets}

\subsubsection{CMU-MMAC} The CMU Multi-Modal Activity Database \cite{de2009guide} contains \textit{indoor} multimodal data of the human activity involving cooking and food preparation. Our research utilizes egocentric video and four of the 3DM-GX1 IMU sensors on left and right arms, and left and right legs. The IMU sample rate is roughly 60 Hz. We follow \cite{nakamura2021sensor} for IMU data preprocessing and get a total of 13 hours of synchronized video and IMUs data, in which 5 hours are annotated with action labels \cite{bansal2022my} of 32 categories.

\subsubsection{WEAR} The WEAR \cite{bock2023wear} dataset is an \textit{outdoor} sports dataset containing visual and IMU data of different workout activities. Our research utilizes egocentric video and all four IMU sensors on left and right wrists, and left and right ankles. The IMU sample rate is 50 Hz. The total duration of the dataset is 15 hours, in which 9 hours are annotated with action labels of 18 categories.

\subsection{Implementation details} 
In this work, we utilize synchronized egocentric video and $N_{imu}=4$ IMU devices attached to the four limbs with a duration of $T=2$ seconds. For video processing, we down-sample and crop the clips to $T_v=16$ frames with a resolution of $H=224, W=224$ pixels. These frames are further divided into $16\times 16$ patches. We apply a masking ratio of $R_v=90\%$, in line with \cite{tong2022videomae}. For the IMU signals, we transform them into spectrograms with a temporal dimension of $T_{imu}=160$ and a frequency dimension of $M_{imu}=128$. These spectrograms are similarly divided into $16\times 16$ patches, with a masking ratio of $R_{\text{imu}} = 75\%$. 
For the network architecture, we adapt the designs for the video and IMU encoders from \cite{tong2022videomae} and \cite{he2022masked}, respectively. The unified encoder consists of a single-layer Transformer. Regarding the training process, we set the loss parameters as follows: $\alpha=1$, $\beta=10$, $\gamma=0.01$, and $\tau=0.05$.

\subsection{Comparison with the state of the art}

As presented in \cref{tab:sota}, we compare our method with the previous state-of-the-art approaches. The pretraining and finetuning data are from the same dataset and all compared methods share the same data for training. Our experiments are differentiated based on modality, including those that use only IMU or video and those that utilize both modalities for training and testing. For the methods that use Transformer \cite{vaswani2017attention} backbones, we all use ViT-Base \cite{dosovitskiy2020vit} for a fair comparison. Below, we present detailed analyses of the results in this table.

\begin{table}[t]
\caption{Comparison with the state of the art. We show action recognition accuracy (mAP$\uparrow$) on the CMU-MMAC~\cite{de2009guide} and WEAR~\cite{bock2023wear} datasets. 
% Our approach is compared with methods that utilize either single or dual modalities, with some requiring pretraining and others not.
}
\label{tab:sota}
\centering
\scalebox{0.8}{
\begin{tblr}{
  cells = {c},
  hline{1-2,6,10,14} = {-}{},
}
                & Modality     & Backbone    & Pretrain     & \scalebox{0.8}{CMU-MMAC (mAP)}  & \scalebox{0.8}{WEAR (Acc)} \\
DCL\cite{ordonez2016deep}           & IMU       & LSTM        & \xmark     & 7.46     & 74.37 \\
ADCL\cite{bock2021improving}           & IMU       & LSTM        & \xmark     & 8.59     & 77.30 \\
LIMU-BERT\cite{xu2021limu}    & IMU       & Transformer & \cmark     & 15.30    & 79.60     \\
Ours      & IMU       & Transformer    & \cmark     & \textbf{31.68}    & \textbf{86.53} \\
SlowFast\cite{feichtenhofer2019slowfast}    & Video     & ResNet50    & \xmark     & 82.03    & 86.13 \\
TimeSformer\cite{bertasius2021space} & Video     & Transformer    & \xmark     & 82.56    & 86.63 \\
VideoMAE\cite{tong2022videomae}    & Video     & Transformer    & \cmark     & 84.63    & 88.47 \\
Ours      & Video     & Transformer    & \cmark     & \textbf{85.07}    & \textbf{89.78} \\
WEAR$+$\cite{bock2023wear}        & IMU-Video & Transformer    & \xmark     & 83.36    & 90.14     \\
CAV-MAE\cite{gong2022contrastive}     & IMU-Image & Transformer    & \cmark     & 74.69    & 90.22     \\
AV-MAE\cite{georgescu2023audiovisual}      & IMU-Video & Transformer    & \cmark     & 84.75    & 91.60     \\
Ours      & IMU-Video & Transformer    & \cmark     & \textbf{87.96}    & \textbf{92.78} 
\end{tblr}
}
\end{table}

% cmu 7028-8428
% cmu 8336-9034
% wear 9994-9014
% wear 9979-9207

\paragraph{Within IMU modality} For training with IMU signals alone (top block of \cref{tab:sota}), we evaluate three prior methods, including DCL \cite{ordonez2016deep} and its improved version ADCL \cite{bock2021improving} that integrates recurrent and convolutional layers to perform supervised pretraining from scratch. Additionally, we evaluate LIMU-BERT \cite{xu2021limu}, which adapts the BERT \cite{devlin2018bert} architecture to process IMU signals, learning general representations from unlabeled IMU data before finetuning with an additional GRU \cite{chung2014empirical} classifier. Compared to these methods, our approach benefits from MAE-based pretraining that learns a stronger IMU representation by reconstructing both raw signals and features. Furthermore, by embedding IMU features within a graph, our graph network can more effectively leverage the relationships between multiple IMUs. Thanks to these advantages, our method enjoys a significant increase (15.30\% $\rightarrow$ 31.68\%) against LIMU-BERT~\cite{xu2021limu} in the IMU-only action recognition accuracy. We further show the effect of each component in the following sections.

\paragraph{Within Video modality} For training exclusively with video data (mid block of \cref{tab:sota}), we assess three methods. SlowFast \cite{feichtenhofer2019slowfast} combines high and low framerate pathways for temporal and spatial semantics with convolutional networks. TimeSformer \cite{bertasius2021space} adapts the Transformer architecture for learning spatiotemporal features from sequences of frame-level patches. In our approach, we design the video branch following the structure proposed in VideoMAE \cite{tong2022videomae}. Our method clearly outperforms SlowFast and TimeSformer, meanwhile slightly surpassing the VideoMAE baseline due to the incorporation of an additional contrastive pretraining objective.

\paragraph{IMU-visual modality} For training that incorporates both modalities (bottom block of \cref{tab:sota}), we evaluate an IMU-video-based method WEAR$+$\cite{bock2023wear} and upgrade its backbone for a fair comparison. We also adapt and evaluate two audio-visual MAE-based pretraining approaches because audio and IMU signals have similar data format. Given that audio data are preprocessed into one-channel spectrograms, we adapt these methods to accept multi-channel IMU spectrograms by adjusting the network's channel numbers. CAV-MAE \cite{gong2022contrastive} processes a spectrogram and a related image as input, and
% . It integrates a contrastive loss with an MAE reconstruction loss during pretraining to investigate the interrelation between the two modalities. 
AV-MAE \cite{georgescu2023audiovisual} utilizes audios and the corresponding videos as inputs.
% , experimenting with various styles of modality fusion. 
Our method surpasses these approaches by employing a graph to model the relationships among multiple collaborative IMU features and by effectively integrating modality fusion during both the pretraining and fine-tuning phases.

\begin{table}[t]
\centering
\caption{Ablation studies. The left table outlines the modality selection, pretraining decision, and the employment of the IMU feature graph. The right tables illustrate the effects of different IMU masking ratios, style options, and pre-training objectives.}
\scalebox{0.8}{
\begin{tblr}{
  column{1} = {c},
  column{2} = {c},
  column{3} = {c},
  column{4} = {c},
  column{5} = {c},
  hline{1-2,6,8,12} = {1-5,7-10}{},
  hline{4-5,9-10} = {7-10}{},
}
\scalebox{0.8}{\textbf{Modality}}  & \scalebox{0.8}{\textbf{Pretrain}}  & \scalebox{0.8}{\textbf{Graph}} & \scalebox{0.8}{CMU-MMAC} & \scalebox{0.8}{WEAR}  &     & \scalebox{0.9}{\textbf{IMU mask ratio}}      & 60\% & 75\% & 90\% \\
IMU       & \xmark     & \xmark     & 24.87    & 79.66 &                    & IMU       & 30.38 & 31.68 & \textbf{31.88}   \\
IMU       & \xmark     & \cmark     & 27.36    & 83.55 &                    & IMU-Video & 83.16 & \textbf{87.96} & 87.62   \\
IMU       & \cmark     & \xmark     & 28.96    & 85.29 &                    &           &   &   &   \\
IMU       & \cmark     & \cmark     & \textbf{31.68}    & \textbf{86.53} &  & \scalebox{0.9}{\textbf{IMU mask style}}       & time & freq. & Both \\
Video     & \xmark     & \xmark     & 69.82    & 86.53 &                    & IMU       & 30.51 & 29.96 & \textbf{31.59}   \\
Video     & \cmark     & \xmark     & \textbf{85.07}    & \textbf{89.78} &  & IMU-Video       & 84.63 & \textbf{87.46} & 86.69   \\
IMU-Video & \xmark     & \xmark     & 71.53    & 90.93 &                    &           &   &   &   \\
IMU-Video & \xmark     & \cmark     & 73.19    & 91.54 &                    & \scalebox{0.9}{\textbf{Supervision}}    & MAE & Contra. & Both \\
IMU-Video & \cmark     & \xmark     & 84.78    & 91.62 &                    & IMU       & 31.45 & 14.32 & \textbf{31.68}   \\
IMU-Video & \cmark     & \cmark     & \textbf{87.96}    & \textbf{92.78} &  & IMU-Video       & 86.63 & 84.65 & \textbf{87.96}   
\end{tblr}}
\label{tab:ablation_main}
\end{table}

\subsection{Ablation studies}

% As shown in \cref{tab:ablation_main}, we perform ablation experiments to evaluate the impact of modality selection, the decision to pre-train, and the incorporation of the IMU feature graph. These experiments, detailed in the left table, demonstrate the effectiveness of our primary contributions. The right tables present additional ablation studies on other critical factors.

\paragraph{Modality correlation and pretraining}
% In \cref{tab:ablation_main}, our experiments demonstrate the effectiveness of pretraining for two main reasons. First, through the process of masking and reconstruction during pretraining, strong representations can be learned, especially by leveraging both modalities, which enables the learning of inter-modal alignment and completion. This experiment also validates our observation that there is strong correlation between the two modalities. Second, in both datasets, the amount of unlabeled data far exceed labeled data. Self-supervised pretraining allows for the full utilization of the entire dataset.
Based on the observation of a strong correlation between ego video and body-worn IMU motion data, we leverage a larger volume of unlabeled data for self-supervised pretraining. As shown in \cref{tab:ablation_main}, our experiments validate the efficacy of pretraining, wherein our model successfully learns mutual alignment and completion between the two modalities.

\paragraph{Collaborative IMUs and feature graph}
% Our experiments show that even without pretraining, the IMU feature graph we proposed can enhance action recognition accuracy. This improvement is due to the graph neural network providing additional information about the relationships among multiple collaborative IMUs during supervised learning, which directs the model's focus to the IMUs relevant to the action and their collaboration. Moreover, incorporating graph-based pretraining further enhances the results, as the model learns how to develop a stronger graph representation during pretraining.
Leveraging the collaborative nature of body-worn IMUs for action capture, we embed their features in a graph, employing graph networks to analyze their relationships. As evidenced in \cref{tab:ablation_main}, using a graph network boosts action recognition accuracy in a supervised context. Additionally, a node masking strategy in pretraining further strengthens our model's capability, enhancing its action recognition efficacy.

% \paragraph{Modality collaboration}
% The experiments show that our method effectively capitalizes on the collaboration between both modalities. Even though videos already contain rich information for action recognition, the IMU provides complementary information that can further enhance accuracy. We present additional visualization results in \cref{sec:vis}.

\paragraph{High masking ratio for both video and IMU} VideoMAE \cite{tong2022videomae} adopts a high masking ratio (90\%) for pretraining, motivated by the significant correlation and redundancy inherent in video data. We find that a high masking ratio (75\%$\sim$90\%) similarly advantages IMU pretraining, because periodic repetitive motions frequently occur in human activities, as illustrated in \cref{fig:final-vis}.

\paragraph{IMU masking style} 
% We find that the IMU masking style does not significantly impact performance. 
We experiment with both structured and unstructured random masking techniques. In structured masking, we implement masking across both temporal and frequency-related patches within the IMU spectrogram. We found that unstructured random masking yields marginally better results.

\paragraph{Pretraining supervision} We discover that incorporating contrastive loss benefits both single-modality and dual-modality pretraining.

\paragraph{Dataset diversity}
\cref{tab:ablation_main} reveals that the accuracy in the IMU modality of the CMU-MMAC dataset is significantly lower compared to the WEAR dataset.
The primary reason is that actions in the CMU-MMAC dataset involve hand-object-environment interactions. However, IMU signals only contain information about hand motion, making it challenging to recognize objects and surroundings.
% In contrast, the WEAR dataset does not involve object or environmental interactions, resulting in higher accuracy.

\begin{table}[t]
    \centering
    \begin{minipage}[b]{0.58\linewidth} % 开始第一个minipage环境
        \centering
        \caption{IMU device missing challenge. Our method experiences a smaller decrease in accuracy compared to others.}
        \scalebox{0.65}{
        \begin{tblr}{
          cells = {c},
          cell{1}{4} = {c=2}{},
          cell{1}{6} = {c=2}{},
          cell{2}{1} = {r=2}{},
          cell{2}{4} = {fg=Gray},
          cell{2}{6} = {fg=Gray},
          cell{3}{5} = {fg=blue},
          cell{3}{7} = {fg=blue},
          cell{4}{1} = {r=2}{},
          cell{4}{4} = {fg=Gray},
          cell{4}{6} = {fg=Gray},
          cell{5}{5} = {fg=blue},
          cell{5}{7} = {fg=blue},
          cell{6}{1} = {r=2}{},
          cell{6}{4} = {fg=Gray},
          cell{6}{6} = {fg=Gray},
          cell{7}{5} = {fg=blue},
          cell{7}{7} = {fg=blue},
          cell{8}{1} = {r=2}{},
          cell{8}{4} = {fg=Gray},
          cell{8}{6} = {fg=Gray},
          cell{9}{5} = {fg=blue},
          cell{9}{7} = {fg=blue},
          hline{1-2,4,6,8,10} = {-}{},
        }
                     & Modality  & Miss  & CMU-MMAC &                     & WEAR  &                    \\
        ADCL\cite{bock2021improving}        & IMU       & \xmark & 8.59     &                     & 77.30 &                    \\
                     & IMU       & \cmark & 5.25     & $[\downarrow 38\%]$ & 70.28 & $[\downarrow 9\%]$ \\
        \scalebox{0.9}{LIMU-BERT\cite{xu2021limu}} & IMU       & \xmark & 15.30    &                     & 79.60 &                    \\
                     & IMU       & \cmark & 11.40    & $[\downarrow 25\%]$ & 73.25 & $[\downarrow 8\%]$ \\
         Ours        & IMU       & \xmark & 31.68    &                     & 86.53 &                    \\
                     & IMU       & \cmark & \textbf{26.47}    & $[\downarrow \textbf{16\%}]$ & \textbf{80.89} & $[\downarrow \textbf{6\%}]$ \\
         Ours        & IMU-Video & \xmark & 87.96    &                     & 92.78 &                    \\
                     & IMU-Video & \cmark & \textbf{85.86}    & $[\downarrow \textbf{3\%}]$  & \textbf{92.42} & $[\downarrow \textbf{1\%}]$
        \end{tblr}
        }

        \label{tab:device-missing}
    \end{minipage}
    \hfill
    \begin{minipage}[b]{0.38\linewidth} % 开始第二个minipage环境
        \centering
        \caption{
        % Feature transferability challenge. We pretrain and finetune in cross dataset settings. 
        % $\{PT\ dataset\} \rightarrow \{FT\ dataset\}$
        Cross dataset pretraining and finetuning. $W$: WEAR, $C$: CMU-MMAC.
        }
        \scalebox{0.65}{
        \begin{tblr}{
          row{1} = {c},
          row{2} = {c},
          row{3} = {c},
          row{4} = {c},
          row{5} = {c},
          row{6} = {c},
          row{7} = {c},
          hline{1-2,8} = {-}{},
        }
        Modality  & Pretrain & $W\rightarrow C$ & $C\rightarrow W$ \\
        IMU       &   \xmark    & 27.36            & 83.55            \\
        IMU       &   \cmark    & 28.08            & 85.27            \\
        Video     &   \xmark    & 69.82            & 86.53            \\
        Video     &   \cmark    & 82.32            & 89.70            \\
        IMU-Video &   \xmark    & 73.19            & 91.54            \\
        IMU-Video &   \cmark    & 85.67            & 91.98            \\
                  &          &                  &                  \\
                  &          &                  &                  
        \end{tblr}
        }
        
        \label{tab:cross-dataset}
    \end{minipage}
\end{table}

\subsection{Performance under real-world challenges}

Since our model employs wearable devices for action recognition, we must account for various challenges in human-centered environments, where individuals might work in diverse environments and have distinct user customizations for their devices. Consequently, we have designed experiments for challenging situations to support more flexible user usages.

\subsubsection{IMU device missing challenge} 

When using wearable devices in various scenarios, the number of IMU sensors required may vary due to budget constraints or physical activity limitations. It is assumed that all IMU devices are available during the pretraining phase. Then, we hypothesize a scenario where a user can only afford half the number of IMUs. While finetuning with fewer devices, a robust pretrained model should continue to exhibit advantages. As shown in \cref{tab:device-missing}, we mask IMU signals from `left wrist' and `left ankle' and finetune the models. Due to the loss of information, a natural decrease in action recognition accuracy is observed. However, our method experiences a less significant drop in accuracy compared to other approaches. Because in our pretraining process, we attempt to reconstruct global information from local data, this approach effectively aligns with this challenge.

\begin{figure}[t]
  \centering
   \includegraphics[width=1.0\linewidth]{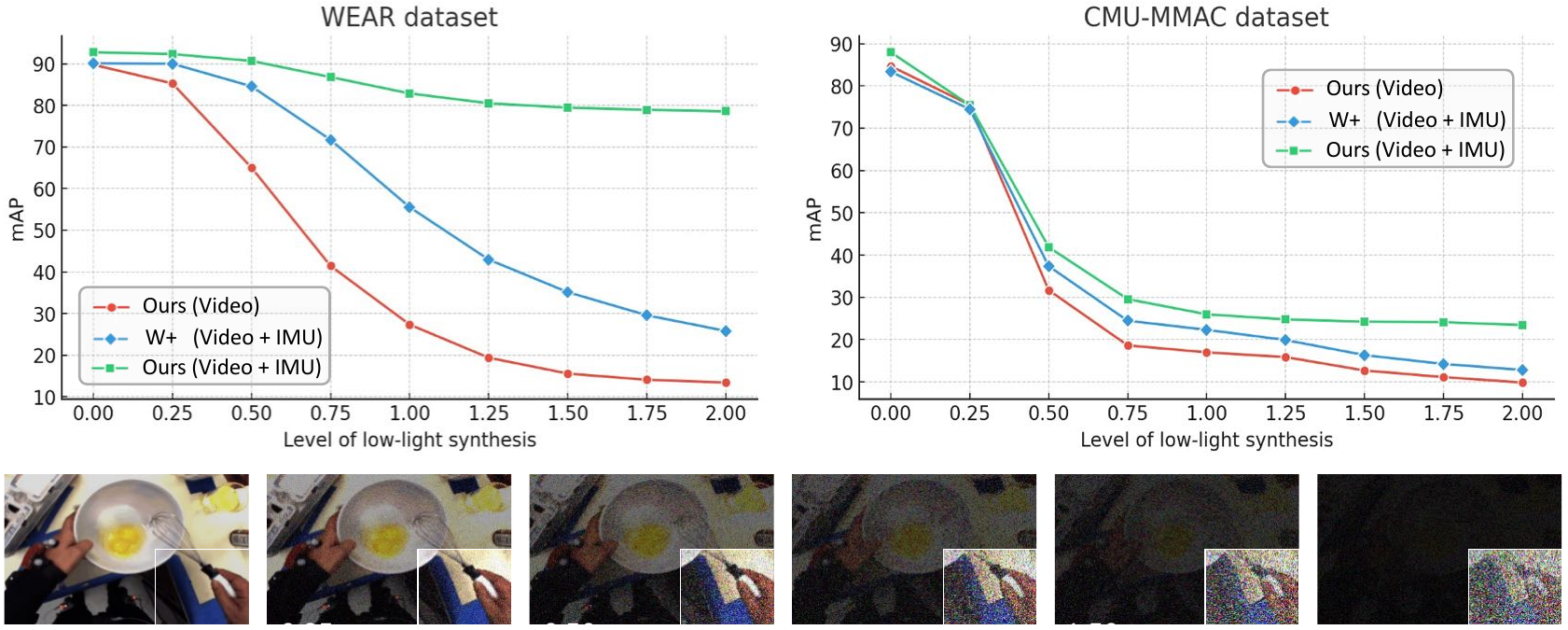}
   % \vspace{-2em}
   \caption{
   Visual degradation challenge. We employ a sophisticated method to synthesize low-light effect and degrade the signal-to-noise ratio on the video input. In such cases, our multimodal model switches focus to IMU modality for robust performance. In contrast, simple multimodal feature concatenation (W+ \cite{bock2023wear}) achieves suboptimal results.
   }
   \label{fig:low-light}
   % \vspace{-1.5em}
\end{figure}

\subsubsection{Feature transferability challenge} 

Datasets that include video and body-worn IMUs may have domain gaps. These gaps arise not only from the potential absence of IMU devices but also from variations in video recording environments and differences in subjects' actions. For IMUs, discrepancies can occur due to variations in sensor accuracy and range. Particularly, the specific location on the body where the device is worn and its orientation can result in distinct differences in the data collected. When finetuning on a dataset that differs from the pretraining dataset, a robust pretrained model should continue to exhibit advantages. We verify this with experiments shown in \cref{tab:cross-dataset} where we pretrain and finetune in cross dataset settings.

\subsubsection{Visual degradation challenge} 

Video models excel at recognizing actions in well-lit environments but struggle in low-light conditions \cite{zheng2020optical}, while motion information from IMU signals naturally resists such variations. However, the challenge remains in determining whether a multimodal model can shift reliance to a more reliable modality for action recognition under low-light conditions. 
Therefore, we devised an experiment in which we degrade the input video quality through a physically-plausible procedure~\cite{wei2020physics,brooks2019unprocessing}, reducing the signal-to-noise ratio and simulating realistic low-light conditions, and then evaluate our multimodal model's performance. As illustrated in \cref{fig:low-light}, our (Video + IMU) model demonstrates robustness across various levels of synthesized low-light scenarios. In contrast, using video only or simply concatenating multimodal features \cite{bock2023wear} achieves suboptimal results. Given that the CMU-MMAC \cite{de2009guide} dataset encompasses actions that involve objects and the surroundings, a noticeable performance decline is naturally observed, since IMU signals hardly contain such information.

\subsection{Qualitative examples}
\label{sec:vis}

\begin{figure}[t]
    \centering
    \includegraphics[width=1.0\linewidth]{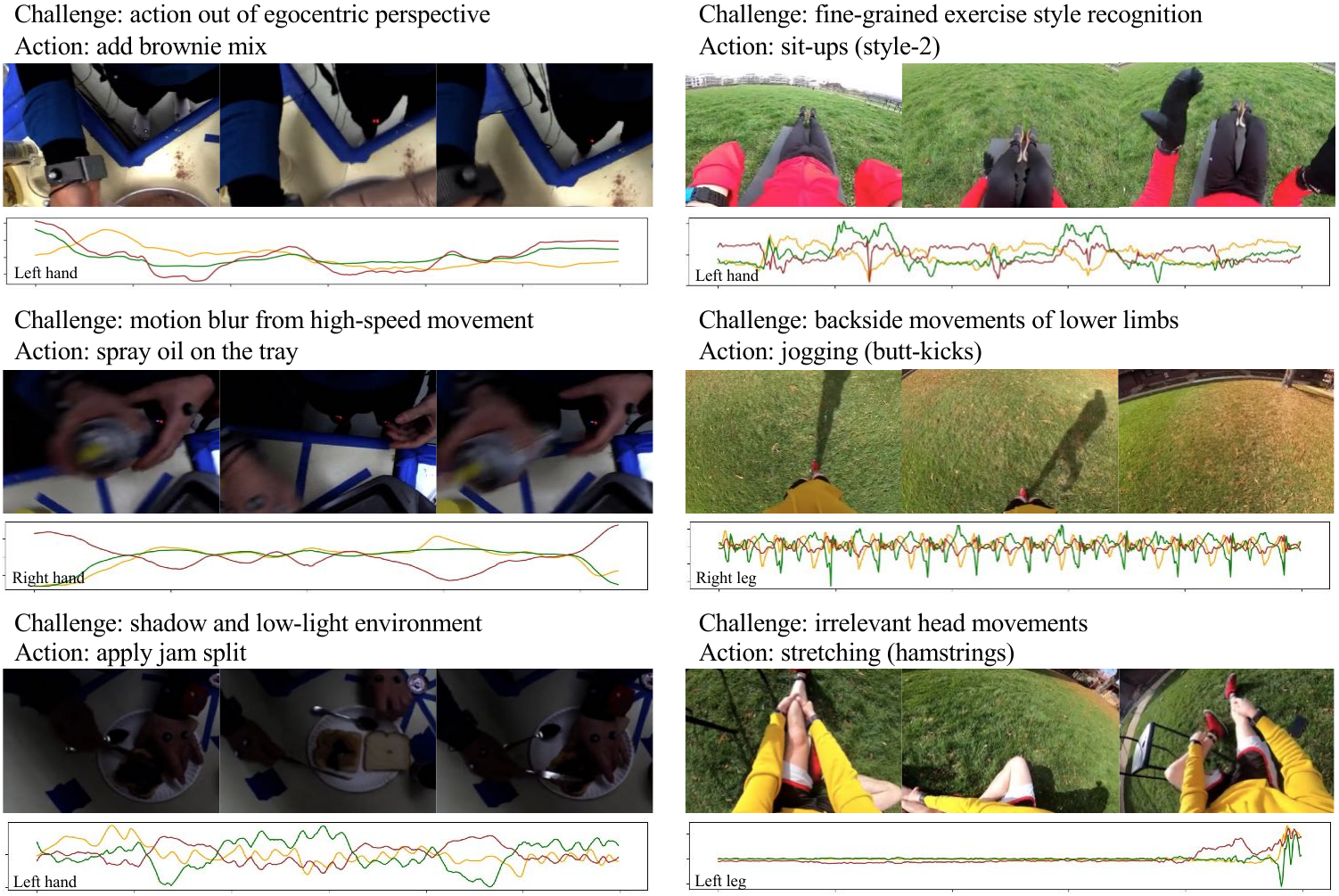}
    \caption{Visualization instances where our multimodal approach successfully recognizes actions, whereas the VideoMAE \cite{tong2022videomae} model failed.}
    \label{fig:final-vis}
\end{figure}

\cref{fig:final-vis} shows several examples of challenging cases where current the state-of-the-art video model \cite{tong2022videomae} struggles to handle. However, our multimodal approach, with the aid of pure limb motion information from IMU that is independent of appearance, demonstrates proficient action recognition capabilities. The left three challenges are derived from the CMU-MMAC \cite{de2009guide} indoor dataset, while the challenges on the right originate from the WEAR \cite{bock2023wear} outdoor dataset.   
\section{Conclusion and Future Directions}

This paper introduces a novel approach to action recognition leveraging egocentric video and IMU sensors, underpinned by an MAE-based framework enhanced with graph neural networks for more holistic visual and motion representation learning. Our method outperforms existing state-of-the-art models in accuracy and robustness across indoor and outdoor datasets. It enables extensive applications in a variety of fields such as professional athletes, workers, and gamers.

Compared with previous IMU-based action recognition methods \cite{ordonez2016deep, bock2021improving}, our method achieves significant accuracy improvement. Nonetheless, it shares the same level of computational demand and inference time as the latest transformer-based MAE methods. Future research could focus on enhancing the computational efficiency.
Another direction is exploring more advanced graph structure designs. We currently utilize a fully connected graph with only four IMU nodes because of the dataset limitation. When more IMUs are available, a more sophisticated graph can be designed to better capture the spatial relations among different body parts.

\bibliographystyle{splncs04}
\bibliography{main}
\end{document}

\title{Supplementary material of\\`Masked Video and Body-worn IMU Autoencoder for Egocentric Action Recognition'}

\titlerunning{Abbreviated paper title}

\author{First Author\inst{1}\orcidlink{0000-1111-2222-3333} \and
Second Author\inst{2,3}\orcidlink{1111-2222-3333-4444} \and
Third Author\inst{3}\orcidlink{2222--3333-4444-5555}}

\authorrunning{F.~Author et al.}

\institute{Princeton University, Princeton NJ 08544, USA \and
Springer Heidelberg, Tiergartenstr.~17, 69121 Heidelberg, Germany
\email{lncs@springer.com}\\
\url{http://www.springer.com/gp/computer-science/lncs} \and
ABC Institute, Rupert-Karls-University Heidelberg, Heidelberg, Germany\\
\email{\{abc,lncs\}@uni-heidelberg.de}}

\maketitle

\section{Implementation details}

\paragraph{Training details} In our experiments, we employ a setup of single NVIDIA A100 GPUs, each with 40GB of memory. This configuration not only supports the computational demands of our tasks but also enhances the reproducibility of our results. For the pretraining phase, our model is trained with a batch size of 24 over 300 epochs, starting with an initial learning rate of 5e-5. This rate undergoes a reduction by a factor of 0.5 every 100 epochs. During the finetuning stage, the batch size is adjusted to 16 for a total of 200 epochs. The learning rate for the encoder is set at 5e-5, with a decay factor of 0.5 applied every 60 epochs. Notably, the learning rate for the MLP classification head is set to be 100 times higher than that of the encoders.

\paragraph{Single-modality pretraining and finetuning} Our multimodal framework is designed to be inherently adaptable, capable of processing inputs from a single modality as demonstrated across most of our experiment results. Specifically, when operating with solely IMU data, our approach offers a robust solution for preserving user privacy. This adaptability is achieved without altering the network's architecture. When only IMU data are utilized, our method involves the pretraining and finetuning of four components: the IMU encoder, the graph encoder, the unified encoder, and the decoders—either pixel and graph decoders or the MLP classification head. When the model operates exclusively on video data, our method involves the video encoder, the unified encoder, and either the pixel decoder or the MLP classification head.

\section{Additional ablation studies}

We carry out additional ablation studies to further validate our model's effectiveness, robustness, and design choices.

The method \cite{gong2022contrastive} based on multimodal MAE for audio and image shows that initializing the pretraining model with weights pretrained on ImageNet consistently enhances model performance. Conversely, VideoMAE \cite{tong2022videomae} reveals that pretraining with image data is not as effective as direct pretraining on video data. 
As illustrated in the left side of \cref{tab:supp}, our findings indicate that while the IMU encoder can be initialized with ImageMAE \cite{he2022masked} trained on ImageNet \cite{deng2009imagenet}, such initialization is not essential for the downstream task of action recognition. Conversely, initializing our video encoder with VideoMAE \cite{tong2022videomae}, trained on Kinetics-400 \cite{kay2017kinetics}, results in a notable improvement in action recognition accuracy. This outcome suggests that IMU data possesses unique characteristics, and that pretraining within the same modality using a larger dataset is beneficial. In our main paper, to ensure a fair comparison, we also initialize previous state-of-the-art methods with VideoMAE trained on Kinetics-400.

In the right side of \cref{tab:supp}, we demonstrate the impact of various graph masking ratios, denoted as $R_g$, during the pretraining phase on downstream action recognition performance. We observe that both excessive (75\%) and insufficient (25\%) masking ratios negatively affect the effectiveness of pretraining. Consequently, we set $R_g=0.5$ to ensure optimal pretraining.

\begin{table}[t]
\centering
\caption{Ablation studies about pretraining network initialization and graph masking ratio $R_g$. Our IMU encoder can be initialized by ImageMAE \cite{he2022masked} trained on ImageNet \cite{deng2009imagenet} and our video encoder can be initialized by VideoMAE \cite{tong2022videomae} trained on Kinetics-400 \cite{kay2017kinetics}.}
\scalebox{0.95}{
\begin{tblr}{
  cells = {c},
  hline{1-2,8} = {1-3,5-7}{},
}
Modality  & Initial. & CMU-MMAC &  & Modality  & $R_g$ & CMU-MMAC \\
IMU       & \xmark              & 31.65    &  & IMU       & 0.25               & 29.42    \\
IMU       & \cmark              & 31.68    &  & IMU       & 0.50               & 31.68    \\
Video     & \xmark              & 77.65    &  & IMU       & 0.75               & 31.45    \\
Video     & \cmark              & 85.07    &  & IMU-Video & 0.25               & 87.33    \\
IMU-Video & \xmark              & 79.89    &  & IMU-Video & 0.50               & 87.96    \\
IMU-Video & \cmark              & 87.96    &  & IMU-Video & 0.75               & 86.26    \\
\end{tblr}
}
\label{tab:supp}
\end{table}

\bibliographystyle{splncs04}
\bibliography{main}